\documentclass[a4paper,twoside]{article}
\usepackage{epsfig}
\usepackage{subcaption}
\usepackage{calc}
\usepackage{amssymb}
\usepackage{amstext}
\usepackage{amsmath}
\usepackage{mathtools}
\usepackage{amsthm}
\usepackage{multicol}
\usepackage{pslatex}
\usepackage{natbib}
\usepackage[bottom]{footmisc}
\usepackage{placeins}
\usepackage{tikz}
\usepackage{amsmath}
\usetikzlibrary{automata}

\usepackage{booktabs}

\usepackage[colorlinks=true, linkcolor=blue, citecolor=blue]{hyperref}
\usepackage[capitalise,noabbrev,nameinlink]{cleveref}

\usepackage{SCITEPRESS}     
\usepackage{algorithm}
\usepackage{algpseudocode}

\usepackage{amsthm}

\theoremstyle{definition}
\newtheorem{definition}{Definition}[section]

\usetikzlibrary{shapes,shadows,arrows,positioning,angles,quotes}
\tikzset{My Arrow Style/.style={single arrow, fill=black!15, anchor=base, align=center,text width=2.3cm}}
\tikzstyle{arrow} = [thick,->,>=stealth]
\usetikzlibrary{calc,trees,positioning,arrows,fit,shapes,calc}

\tikzstyle{startstop} = [rectangle, rounded corners, minimum width=1.5cm, minimum height=0.5cm,text centered, draw=black, fill=red!30]
\tikzstyle{io} = [trapezium, trapezium left angle=70, trapezium right angle=110, minimum width=1cm, minimum height=0.5cm, text centered, draw=black, fill=blue!30]

\tikzstyle{process} = [rectangle, minimum width=3cm, minimum height=0.5cm, text centered, draw=black, fill=orange!30]
\tikzstyle{process2} = [rectangle, minimum width=1cm, minimum height=0.5cm, text centered, draw=black, fill=orange!30]
\tikzstyle{decision} = [diamond, minimum width=1cm, minimum height=0.5cm, text centered, draw=black, fill=green!30]

\tikzstyle{arrow} = [thick,->,>=stealth]

\newcommand{\scaleTable}[1]{\scalebox{0.795}{#1}}

\begin{document}

\title{Turn-based Multi-Agent Reinforcement Learning Model Checking}

\author{\authorname{Dennis Gross}
\affiliation{Institute for Computing and Information Sciences, Radboud University, Toernooiveld 212, 6525 EC Nijmegen, \\The Netherlands}
\email{dgross@science.ru.nl}
}


\abstract{
In this paper, we propose a novel approach for verifying the compliance of turn-based multi-agent reinforcement learning (TMARL) agents with complex requirements in stochastic multiplayer games.
Our method overcomes the limitations of existing verification approaches, which are inadequate for dealing with TMARL agents and not scalable to large games with multiple agents.
Our approach relies on tight integration of TMARL and a verification technique referred to as model checking. 
We demonstrate the effectiveness and scalability of our technique through experiments in different types of environments.
Our experiments show that our method is suited to verify TMARL agents and scales better than naive monolithic model checking.}

\onecolumn \maketitle \normalsize \setcounter{footnote}{0} \vfill

\section{\uppercase{Introduction}}
AI technology has revolutionized the game industry~\citep{DBLP:journals/corr/abs-1912-06680}, enabling the creation of agents that can outperform human players using \emph{turn-based multi-agent reinforcement learning (TMARL)}~\citep{DBLP:journals/nature/SilverHMGSDSAPL16}.
TMARL consists of multiple agents, where each one learns a near-optimal policy based on its own objective by making observations and gaining rewards through turn-based interactions with the environment~\citep{wong2022deep}.

The strength of these agents can also be a problem, limiting the gameplay experience and hindering the design of high-quality games with non-player characters (NPCs)~\citep{DBLP:journals/gamestudies/Svelch20a,DBLP:journals/tciaig/NamHI22}.
Game developers want to ensure that their TMARL agents behave as intended, and tracking their rewards can allow them to fine-tune their performance.
However, rewards are not expressive enough to encode more complex requirements for TMARL agents, such as ensuring that a specific sequence of events occurs in a particular order~\citep{littman2017environment,DBLP:conf/tacas/HahnPSSTW19,DBLP:conf/formats/HasanbeigKA20,DBLP:journals/aamas/VamplewSKRRRHHM22}.

This paper addresses the challenge of verifying the compliance of TMARL agents with complex requirements by combining TMARL with \emph{rigorous model checking} \citep{DBLP:books/daglib/0020348}.
Rigorous model checking is a formal verification technique that uses mathematical models to verify the correctness of a system with respect to a given property. It is called "rigorous" because it provides guarantees of correctness based on rigorous mathematical reasoning and logical deductions.
In the context of this paper, rigorous model checking is used to verify TMARL agents.
The system being verified is the TMARL system, which is modeled as a \emph{Markov decision process (MDP)} treating the collection of agents as a \emph{joint agent}, and the property is the set of requirements that the agents must satisfy.
Our proposed method\footnote{GitHub-Repository: \url{https://github.com/DennisGross/COOL-MC/tree/markov_games}} supports a broad range of properties that can be expressed by \emph{probabilistic computation tree logic} \citep[PCTL;][]{DBLP:journals/fac/HanssonJ94}.
We evaluate our method on different TMARL benchmarks and show that it outperforms \emph{naive monolithic model checking}\footnote{Naive monolithic model checking is called "naive" because it does not take into account the complexity of the system or the number of possible states it can be in, and it is called "monolithic" because it treats the entire system as a single entity, without considering the individual components of the system or the interactions between them.}.

\noindent To summarize, the \textbf{main contributions} of this paper~are:
\begin{enumerate}
    \item rigorous model checking of TMARL agents,
    \item a method that outperforms naive monolithic model checking on different benchmarks.
\end{enumerate}
The paper is structured in the following way.
First, we summarize the related work and position our paper in it.
Second, we explain the fundamentals of our technique.
Then, we present the TMARL model checking method and describe its functionalities and limitations.
After that, we evaluate our method in multiple environments from the AI and model checking community~\citep{DBLP:conf/cig/LeeT17,even2006action,abu2019tic,DBLP:conf/tacas/HartmannsKPQR19}.
The empirical analysis shows that the TMARL model checking method can effectively check PCTL properties of TMARL agents.
\section{\uppercase{Related Work}}

\emph{PRISM}~\citep{DBLP:conf/cav/KwiatkowskaNP11} and \emph{Storm}~\citep{DBLP:journals/sttt/HenselJKQV22} are tools for formal modeling and analysis of systems that exhibit uncertain behavior.
PRISM is also a language for modeling discrete-time Markov chains (DTMCs) and MDPs.
We use PRISM to model the TMARL environments as MDPs.
Until now, PRISM and Storm do not allow verifying TMARL agents.
PRISM-games~\citep{DBLP:journals/sttt/KwiatkowskaPW18} is an extension of PRISM to verify stochastic multi-player games (including turn-based stochastic multi-player games).
Various works about turn-based stochastic game model checking have been published~\citep{DBLP:journals/corr/abs-2211-06141,DBLP:conf/birthday/KwiatkowskaN019,DBLP:conf/uai/LiZR020,DBLP:journals/jacm/HansenMZ13,DBLP:books/cu/11/000111a}.
None of them focus on TMARL systems.
TMARL has been applied to multiple turn-based games~\citep{DBLP:conf/cig/WenderW08,DBLP:conf/iva/PagalyteMC20,DBLP:journals/nature/SilverHMGSDSAPL16,DBLP:journals/aai/VidegainG21,DBLP:conf/iva/PagalyteMC20}.
The major work about model checking for RL agents focuses on single RL agents~\citep{yuwangPCTL,DBLP:conf/formats/HasanbeigKA20,DBLP:conf/tacas/HahnPSSTW19,DBLP:conf/aiia/HasanbeigKA19,fulton2019verifiably,DBLP:conf/cdc/SadighKCSS14,bouton2019reinforcement,DBLP:conf/aaai/Chatterjee0PRZ17}.
However, model checking work exists for cooperative MARL~\citep{riley2021reinforcement,DBLP:conf/iros/KhanZLWSTRBK19,DBLP:conf/kes/RileyCPKB21}, but no work for TMARL.
Therefore, with our research, we try to close the gap between TMARL and model checking.

\section{\uppercase{Background}}

In this section, we introduce the fundamentals of our work. We begin by summarizing the modeling and analysis of probabilistic systems, which forms the basis of our approach to check TMARL agents. We then describe TMARL in more detail.

\subsection{Probabilistic Systems}
A \textit{probability distribution} over a set $X$ is a function $\mu : X \rightarrow [0,1]$ with $\sum_{x \in X} \mu(x) = 1$. The set of all distributions on $X$ is denoted $Distr(X)$.

\begin{definition}[Markov Decision Process]
A \emph{Markov decision process (MDP)} is a tuple $M = (S,s_0,A,T,rew)$ where
$S$ is a finite, nonempty set of states; $s_0 \in S$ is an initial state; $A$ is a finite set of actions; $T\colon S \times A \rightarrow Distr(S)$ is a probability transition function;
$rew \colon S \times A \rightarrow \mathbb{R}$ is a reward~function.
\end{definition}
We employ a factored state representation where each state $s$ is a vector of features $(f_1, f_2, ...,f_n)$ where each feature $f_j \in \mathbb{Z}$ for $1 \leq i \leq n$ ($n$ is the dimension of the state).
The available actions in $s \in S$ are $A(s) = \{a \in A \mid T(s,a) \neq \bot\}$.
An MDP with only one action per state ($\forall s \in S : |A(s)| = 1$) is a DTMC.
A path of an MDP $M$ is an (in)finite sequence $\tau = s_0 \xrightarrow[\text{}]{\text{$a_0,r_0$}} s_1 \xrightarrow[\text{}]{\text{$a_1,r_1$}}...$, where $s_i \in S$, $a_i \in A(s_i)$, $r_i \vcentcolon= rew(s_i,a_i)$, and $T(s_i,a_i)(s_{i+1}) \neq 0$.
A state $s'$ is reachable from state $s$ if there exists a path $\tau$ from state $s$ to state $s'$.
We say a state $s$ is reachable if $s$ is reachable from $s_0$.

\begin{definition}[Policy]
A \emph{memoryless deterministic policy} for an MDP $M$ is a function $\pi \colon S \rightarrow A$ that maps a state $s \in S$ to an action $a \in A(s)$.
\end{definition}

Applying a policy $\pi$ to an MDP $M$ yields an \emph{induced DTMC}, denoted as $D$, where all non-determinism is resolved.
A state $s$ is reachable by a policy $\pi$ if $s$ is reachable in the DTMC induced by $\pi$.
We specify the properties of a DTMC via the specification language PCTL~\citep{yuwangPCTL}.
\begin{definition}[PCTL Syntax]\label{def:pctl}
    Let $AP$ be a set of atomic propositions. The following grammar defines a state formula:
    $\Phi  ::= \text{ true }|\text{ a }| \text{ } \Phi_1 \land \Phi_2 \text{ }|\text{ }\lnot \Phi\text{ }|P_{\bowtie p}| P^{max}_{\bowtie p}(\phi)\text{ }|\text{ }P^{min}_{\bowtie p}(\phi)$ where $a \in AP, \bowtie \in \{<,>,\leq,\geq\}$, $p \in [0,1]$ is a threshold, and $\phi$ is a path formula which is formed according to the following grammar $\phi ::= X\Phi\text{ }|\text{ }\phi_1\text{ }U\text{ }\phi_2\text{ }|\text{ }\phi_1\text{ }F_{\theta t}\text{ }\phi_2\text{ }|G\text{ }\Phi$ with $\theta = \{<,\leq\}$.
\end{definition}

For MDPs, PCTL formulae are interpreted over the states of the induced DTMC of an MDP and a policy.
In a slight abuse of notation, we use PCTL state formulas to denote probability values. That is, we sometimes write $P_{\bowtie p}(\phi)$ where we omit the threshold $p$. 
For instance, in this paper, $P(F\text{ }collision)$  denotes the reachability probability of eventually running into a collision.
There exist a variety of model checking algorithms for verifying PCTL properties~\citep{DBLP:conf/focs/CourcoubetisY88,DBLP:journals/jacm/CourcoubetisY95}.
PRISM~\citep{DBLP:conf/cav/KwiatkowskaNP11} and Storm~\citep{DBLP:journals/sttt/HenselJKQV22} offer efficient and mature tool support for verifying probabilistic systems~\citep{DBLP:conf/cav/KwiatkowskaNP11,DBLP:journals/sttt/HenselJKQV22}.

\begin{definition}[Turn-based stochastic multi-player game]
    A \emph{turn-based stochastic multi-player game (TSG)} is a tuple $(S, s_0, I, A, (S_i)_{i \in I}, T, \{rew_i\}_{i \in I})$ where $S$ is a finite, nonempty set of states; $s_0 \in S$ is an initial state; $I$ is a finite, nonempty set of agents; $A$ is a finite, nonempty set of actions available to all agents; $(S_i)_{i \in I}$ is a partition of the state space $S$; $T \colon S \times A  \rightarrow [0,1]$ is a transition function; and $rew_i \colon S_i  \times A \rightarrow \mathbb{R}$ is a reward function for each agent $i$. 
\end{definition}

Each agent $i \in I$ has a policy $\pi_i \colon S_i \rightarrow A$ that maps a state $s_i \in S_i$ to an action $a_i \in A$.
The \emph{joint policy}~$\pi$ induced by the set of agent policies $\{\pi_i\}_{i \in I}$ is the mapping from states into actions and transforms the TSG into an induced DTMC.

\subsection{Turn-based Multi-Agent Reinforcement Learning (TMARL)}
\begin{figure}[tbp]
\centering
\scalebox{0.795}{
    
    \begin{tikzpicture}[]
     {};
    \node (env) [process, below of=agent1,yshift=-0.25cm,xshift=2cm] {Environment};
    \node (agent1) [process] {Agent 1 $\pi_1(s)$};
    \node (agent2) [process, below of=env,yshift=-0.25cm,xshift=2cm] {Agent 2 $\pi_2(s)$};
    
    \draw [arrow] (agent1) -| node[anchor=west] {$a_1$} (env);
    \draw [arrow] (env) -| node[anchor=east] {$s_1,r_1$} (agent1);

    \draw [arrow] (agent2) |- node[anchor=west] {$a_2$} (env);
    \draw [arrow] (env) |- node[anchor=east] {$s_2,r_2$} (agent2);
    \end{tikzpicture}
}
\caption{This diagram represents a TMARL system in which two agents (Agent 1 and Agent 2) interact in a turn-based manner with a shared environment. The agents receive states (denoted as $s_1$ and $s_2$) and rewards (denoted as $r_1$ and $r_2$) from the environment based on their previous actions (denoted as $a_1$ and $a_2$). The agents then use this information to select their next actions, which they send back to the environment.}
\label{fig:tmarl}
\end{figure}
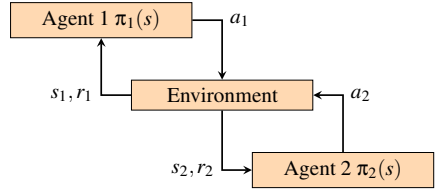

We now introduce TMARL.
The standard learning goal for RL is to find a policy $\pi$ in an MDP such that $\pi$ maximizes the expected discounted reward, that is, $\mathbb{E}[\sum^{L}_{t=0}\gamma^t R_t]$, where $\gamma$ with $0 \leq \gamma \leq 1$ is the discount factor, $R_t$ is the reward at time $t$, and $L$ is the total number of steps~\citep{kaelbling1996reinforcement}.
TMARL extends the RL idea to find near-optimal agent policies $\pi_i$ in a TSG setting (compare \cref{fig:rl} with \cref{fig:tmarl}). 
Each policy $\pi_i$ is represented by a neural network.
A neural network is a function parameterized by weights $\theta_i$.
The neural network policy $\pi_i$ can be trained by minimizing a sequence of loss functions~$J(\theta_i,s,a_i)$~\citep{mnih2013playing}.

\begin{figure}[tbp]
\centering
\scalebox{0.795}{
    \begin{tikzpicture}[]
     {};
    \node (agent1) [process] {Agent 1 $\pi(s)$};
    \node (env) [process, below of=agent1,yshift=-0.25cm,xshift=2cm] {Environment};
    
    \draw [arrow] (agent1) -| node[anchor=west] {$a$} (env);
    \draw [arrow] (env) -| node[anchor=east] {$s,r$} (agent1);
    \end{tikzpicture}
}
\caption{This diagram represents a single RL system in which an agent (Agent 1) interacts with an environment. The agent observes a state (denoted as $s$) and a reward (denoted as $r$) from the environment based on its previous action (denoted as $a$). The agent then uses this information to select the next action, which it sends back to the environment.}
\label{fig:rl}
\end{figure}
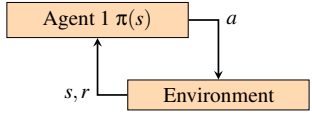

\section{Model Checking of TMARL agents}
We now describe how to verify trained TMARL agents.
Recall, the joint policy $\pi$ induced by the set of all agent policies $\{\pi_i\}_{i \in I}$ is a single policy $\pi$.
The tool \emph{COOL-MC}~\citep{DBLP:conf/setta/Gross22} allows model checking of a single RL policy $\pi$ against
a user-provided PCTL property $P(\phi)$ and MDP $M$. Thereby, it builds the induced DTMC $D$ incrementally~\citep{DBLP:conf/concur/CassezDFLL05}.

\begin{figure}[tbp]
\centering
\scalebox{0.795}{

    \begin{tikzpicture}[]
     {};
    \node (input_state) [io] {State $s$};
    \node (preprocess) [process, below of=input_state] {Extract $turn$ from state $s$};
    \node (dec1) [decision, below of=preprocess,yshift=-1cm] {Which $turn$?};
    \node (agent1) [process2, left of=dec1,xshift=-2.2cm] {$\pi_1(s)$};
    \node (agent2) [process2, right of=dec1,xshift=2.2cm] {$\pi_2(s)$};
    \node (output1) [io, below of=dec1, yshift=-1cm] {Action $a$};

    \draw [arrow] (input_state) -- node[anchor=west] {} (preprocess);
    \draw [arrow] (preprocess) -- node[anchor=west] {} (dec1);
    \draw [arrow] (dec1) -- node[anchor=south] {$turn=1$} (agent1);
    \draw [arrow] (dec1) -- node[anchor=south] {$turn=2$} (agent2);

    \draw [arrow] (agent1) |- node[anchor=east] {$a_1$} (output1);
    \draw [arrow] (agent2) |- node[anchor=west] {$a_2$} (output1);

    \node [draw, dashed, minimum width=7.6cm, minimum height=4cm,xshift=0.0cm,yshift=-2.5cm]{};
    \node[] at (-2.38,-0.3) {\textit{Joint Policy Wrapper}};
    \end{tikzpicture}
}
\caption{An example of a joint policy wrapper with two policies. The wrapper takes in a state (denoted as $s$) and extracts the current turn from that state. It then uses this information to determine which of two policies ($\pi_1$ and $\pi_2$) should choose the next action. The selected policy then produces an action, which is output by the joint policy wrapper.}
\label{fig:joint_agent_wrapper}
\end{figure}
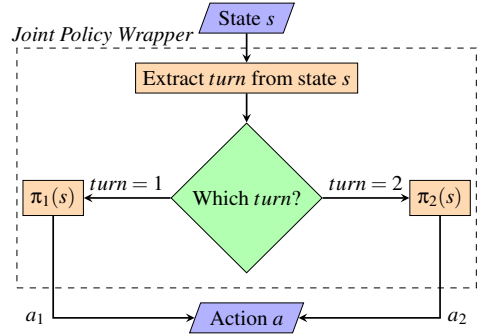

To verify a TMARL system, we model it as a normal MDP.
We have to extend the MDP with an additional feature called \emph{turn} that controls which agent's turn it is.
To support joint policies $\pi(s)$, and therefore multiple TMARL agents, we created a \emph{joint policy wrapper} that queries the corresponding TMARL agent policy at every turn (see \cref{fig:joint_agent_wrapper}).
With the joint policy wrapper, we build the induced DTMC the following way.
For every state $s$ that is reachable via the joint policy $\pi$, we query for an action $a = \pi(s)$.
In the underlying MDP $M$, only states $s'$ that may be reached via that action $a \in A(s)$ are expanded.
The resulting DTMC induced by $M$ and $\pi$ is fully deterministic, as no action choices are left open and ready for efficient model checking.

\paragraph{Limitations.}
Our method allows the model checking of probabilistic policies by always choosing the action with the highest probability at each state. We support any environment that can be modeled using the PRISM language~\citep{DBLP:conf/cav/KwiatkowskaNP11}. However, our method does not consider PCTL properties with the reward operator~\citep{DBLP:journals/fac/HanssonJ94}. When creating the joint policy, there is no separation of which agent receives which reward.

TSGs with more than two agents must handle inactive agents who no longer participate in the game. Our method handles this by allowing non-participating agents to only apply actions that only change the turn feature, allowing the next agent to make a move. This must be considered when using the \emph{expected time step PCTL operator}~\cite {DBLP:journals/fac/HanssonJ94}.

Our method is independent of the learning algorithm and allows for the model checking of TMARL policies that select their actions based on the current and fully-observed state. For simplicity, we focus on TMARL agents with the same action space in this paper. However, extending our method to support TMARL agents with different action spaces and different partial observations for different agents is~straightforward.


\section{\uppercase{Experiments}}
We now evaluate our proposed model checking method in multiple environments.

\subsection{Setup}
In this section, we provide an overview of the experimental setup. We first introduce the environments, followed by the trained TMARL agents. Next, we describe the model checking properties that we used, and finally, we provide details about the technical~setup.
\begin{table*}[tbp]
\centering
\scaleTable{
\begin{tabular}{@{}lllrrrr@{}}
\toprule
\textbf{Env.}     & \textbf{Label}                         & \textbf{PCTL Property Query ($P(\phi)$)}                                                                             &  \textbf{$=$}       & \textbf{$|S|$}        & \textbf{$|T|$}    & \textbf{Time (s)}                                                                              \\ \midrule 
Pokemon                & won1                             & $P(F\text{ }won_1)$                                                                            &  $TO$ &   $TO$ &  $TO$  & $TO$   \\ \midrule
Pokemon (5HP)                & HP5NoHealP1                             & $P(F\text{ }won_1)$                                                                            &  $0.34$ &   $213$ &  $640$  & $14$   \\ \midrule
Pokemon (5HP)                & HP5NoHealP2                             & $P(F\text{ }won_2)$                                                                            &  $0.66$ &   $222$ &  $667$  & $14$ \\ \midrule
Pokemon (5HP)                & usePoisons0HealP1                             & $P(poisons_1=2 \text{ U } poisons_1<2)$                                                                            &  $0.0$ &   $238$ &  $715$  & $17$ \\ \midrule
Pokemon (20HP)                & useHeal20P1                             & $P(healpot_1=1\text{ U }healpot_1=0)$                                                                            &  $0.65$ &   $6720$ &  $40315$  & $401$   \\ \midrule
MABP 25               & lost1                             & $P(\text{F }lost_1)$                                                                            &  $0.0$ &   $50$ &  $51$  & $5$   \\ \midrule
MABP 100               & lost1                             & $P(\text{F }lost_1)$                                                                            &  $0.0$ &   $100$ &  $100$  & $296$   \\ \midrule
Tic-Tac-Toe             & marking\_order                             & $P(((cell_{10}=0 \text{ U } cell_{10}=2) \text{ U } cell_{12}=2) \text{ U } cell_{11}=2)$                                                                            &  $1.0$ &   $18$ &  $30$  & $0.5$   \\ \midrule
CC            & CC1KO                           & $P(F\text{ }player1\_ko)$                                                                            &  $0.0$ &   $205$ &  $281$  & $54$   \\ \midrule
CC            & CC2KO                           & $P(F\text{ }player2\_ko)$                                                                            &  $0.001$ &   $197$ &  $273$  & $53$   \\ \midrule
CC            & CC3KO                           & $P(F\text{ }player3\_ko)$                                                                            &  $0.0$ &   $205$ &  $281$  & $54$   \\ \midrule
CC            & collision                           & $P(F\text{ }collision)$                                                                            &  $0.36$ &   $199$ &  $275$  & $47$   \\ \bottomrule

\end{tabular}
}
\caption{
The table presents the results of running probabilistic model checking via our method on various environments. For each environment, the table lists the label for the probabilistic computation tree logic (PCTL) property query that was used, the result of the query, the number of states in the environment, the number of transitions, and the time it took to run the query. $TO$ indicates that the query did not complete within 24 hours, and therefore the time taken is unknown.}
\label{tab:pctl_labels}
\end{table*}

\paragraph{Environments.}
\begin{figure}[tbp]
  \centering
   {\epsfig{file = 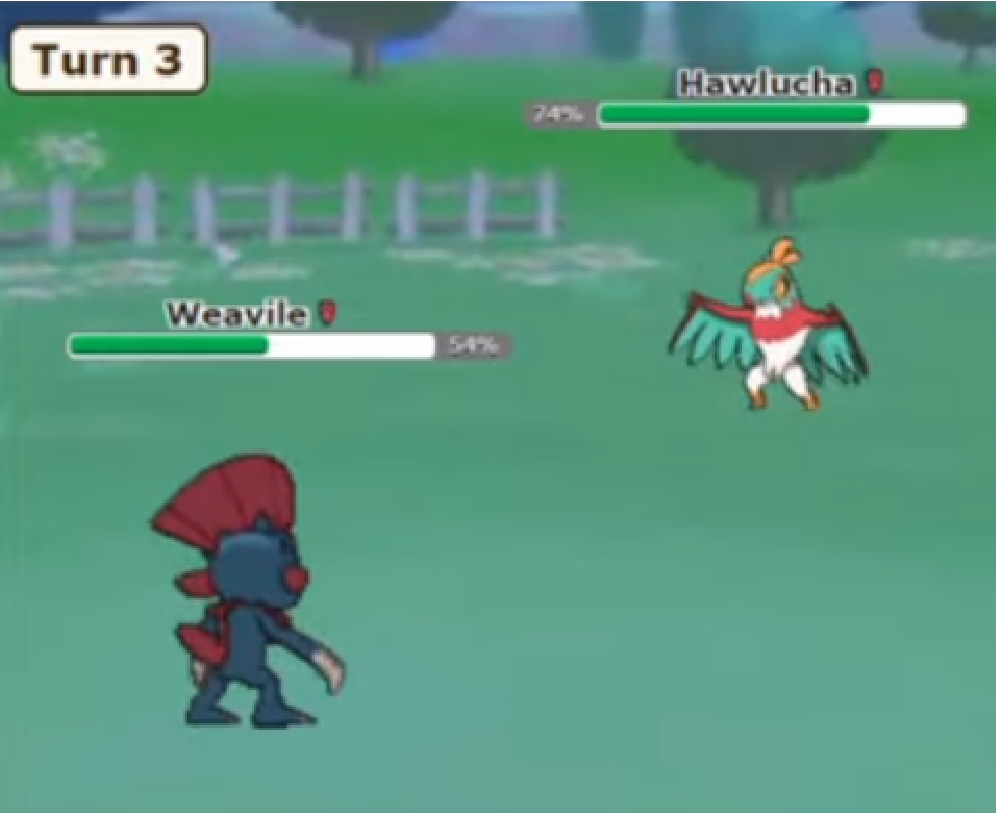, width = 7.2cm}}
  \caption{This screenshot shows a scene from the Showdown AI competition, in which two Pokemon characters are engaged in a battle. We model this scene in PRISM. The AI-controlled Pokemons use different policies $\pi_i$ to try and defeat their opponent. The outcome of the battle will depend on the abilities and actions of the two Pokemons, as well as on random elements.}
  \label{fig:pokemon}
\end{figure}

\emph{Pokemon} is an environment from the game franchise Pokemon that was developed by Game
Freak and published by Nintendo in 1996~\citep{freak1996pokemon}.
It is used in the Showdown AI competition \citep{DBLP:conf/cig/LeeT17}.
In a Pokemon battle, two agents
fight one another until the Pokemon of one agent is knocked out (see \cref{fig:pokemon}).
The impact of randomness in Pokemon is significant, and much of the game's competitive strategy comes from accurately estimating stochastic events.
The damage calculation formula for attacks includes a random multiplier between the
values of 0.85 and 1.0.
Each Pokemon has four different attack actions (tackle, punch, poison, sleep) and can use items (for example, heal pots to recover its hit points (HP)).
The attacks \emph{tackle} and \emph{punch} decrease the opponent's Pokemon HP, \emph{poison} decreases the HP over multiple turns, and \emph{sleep} does not allow the opponent's Pokemon to attack for multiple turns. 
All actions in Pokemon have a success rate, and there is a chance that they fail.
\begin{gather*}
S = \{(turn,done,HP_0,sleeping_0,poisoned_0,\\healpots_0,sleeps_0,poisons_0,punches_0,\\HP_1,sleeping_1,poisoned_1,healpots_1,sleeps_1,\\poisons_1,punches_1),...\} \\
Act = \{sleep, tackle, heal,punch,poison\}\\
rew_i = \begin{cases}
        \text{if agent } i \text{ wins:} \\
        5000\\+ max(100-HP_1-0.2 *(100-HP_0),0)
        \\
        \text {otherwise:}\\
        max(100-HP_1-0.2 *(100-HP_0),0)\\
        \end{cases}
\end{gather*}
The main purpose of this environment is to show that it is possible to verify TMARL agents in complex environments.
The main difference to the Showdown AI competition is that each agent observes the full game state, each agent has only the previously mentioned action choices, and our environment allows one Pokemon per agent.

The \emph{multi-armed bandit problem (MABP)} is a problem in which a fixed limited set of resources must be allocated between competing choices in a way that maximizes their expected gain when each choice's properties are only partially known at the time of allocation and may become better understood as time passes or by allocating resources to the choice~\citep{even2006action,DBLP:conf/icassp/ShahrampourRJ17}. It is a classic RL problem. We transformed this problem into a turn-based MABP.
At each turn, an agent has to learn which action maximizes its expected reward.
\begin{gather*}
S = \{(HP_1,HP_2,...,HP_N, turn, done),...\} \\
Act = \{bandit_1,bandit_2\}\\
rew_i = \begin{cases}
        1 \text{, if agent $i$ is alive}
        \\
        0 \text {, otherwise}
        \end{cases}
\end{gather*}

The \emph{Tic-Tac-Toe} environment is a paper-and-pencil game for two agents who take turns marking the empty cells in a 3x3 grid with X or O. The agent who succeeds in placing three of their marks in a horizontal, vertical, or diagonal row is the winner~\citep{abu2019tic}.
With a probability of $10\%$, an agent does not draw in the grid during its~turn.

\begin{gather*}
S = \{(cell_{00}, cell_{01}, cell_{02},cell_{10}, cell_{11}, cell_{12},\\cell_{20}, cell_{21}, cell_{22}, turn, done),...\} \\
Act = \text{A mark action per cell.}\\
rew_i = \begin{cases}
        500\text{, if agent $i$ wins}\\ 
        0\text{, otherwise}\\
        \end{cases}
\end{gather*}
The \emph{Coin Collection (CC)} environment is a game in which three agents must collect coins in a 4x4 grid world without colliding with each other. If an agent collides with another, the environment terminates.
An agent can attack another agent by standing next to them with a success rate of $0.4$. Each agent receives a reward for every round it is not knocked out (its HP is not 0) and a larger reward for collecting coins. The CC environment is inspired by the QComp benchmark resource gathering~\citep{DBLP:conf/tacas/HartmannsKPQR19}.
\begin{gather*}
S = \{(x_1,y_1,hp_1,x_2,y_2,hp_2,x_3,y_3,hp_3,\\coin_x,coin_y,done,turn),...\} \\
Act = \{up,right,down,left,\\hit\_up,hit\_right,hit\_down,hit\_left\}\\
rew_i = \begin{cases}
        100\text{, if agent $i$ collects coin} \\ 
        1\text{, otherwise}\\
        \end{cases}
\end{gather*}

\paragraph{Trained TMARL agents.}
In the training results, agent 1 in Pokemon has an average reward of $818.23$ over 100 episodes, while agent 2 has an average reward of $690.32$ over the same number of episodes ($50,000$ episodes in total). In Tic-Tac-Toe ($10,000$ episodes in total), agent 1 has an average reward of $370.0$, while agent 2 has an average reward of $100$.
In CC ($10,000$ episodes in total), agent 1 has an average reward of $29.72$, agent 2 has an average reward of $111.58$, and agent 3 has an average reward of $117.69$.
The reward of the TMARL agents can be neglected because we only use them for performance measurements.
All of our training runs used a seed of $128$, an $\epsilon=0.5$. $\epsilon_{min}=0.1$, $\epsilon_{dec}=0.9999$, $\gamma=0.99$, a learning rate of $0.0001$, batch size of $32$, replay buffer size of $300,000$, and a target network replacement interval of $304$.
The Pokemon agents have four layers, each with 256 rectifier~neurons.
The Tic-Tac-Toe, MABP, and CC agents have two layers, each with 256 rectifier~neurons.

\paragraph{Properties.}
\cref{tab:pctl_labels} presents the property queries of the trained policies.
For example, $HP5NoHealP1$ describes the probability of agent 2 winning the Pokemon battle when both Pokemon have $HP=5$ and no more heal pots.
Note, at this point, that our main goal is to verify the trained TMARL policies and that we do not focus on training near-optimal policies.

\paragraph{Technical setup.}
We executed our benchmarks on an NVIDIA GeForce GTX 1060 Mobile GPU, 16 GB RAM, and an Intel(R) Core(TM) i7-8750H CPU~@~2.20GHz~x~12.
For model checking, we use Storm 1.7.1 (dev).

\subsection{Analysis}
In this section, we address the following research questions:
\begin{enumerate}
    \item Does our proposed method scale better than naive monolithic model checking?
    \item How many TMARL agents can our method handle?
    \item Do the TMARL agents perform specific game moves?
\end{enumerate}
We will provide detailed answers to these questions and discuss the implications of our findings.

\paragraph{Does our proposed method scale better than naive monolithic model checking?}
In this experiment, we compare our method with a naive monolithic model checking in the Pokemon environment.
For a whole Pokemon battle (starting with HP=100, unlimited tackle attacks, 5 punch attacks, 2 sleep attacks, 2 poison attacks, and 3 heal pots), naive monolithic model checking runs out of memory.
On the other hand, our method runs out of time (time out after 24 hours).
However, we can train TMARL agents in the Pokemon environment and can, for example, analyze the end game.
For instance, our method allows the model checking of environments with HP of 20 and 1 heal pot left, and we can quantify the probability that Pokemon 1 uses the heal pot with $0.65$ (see useHeal20P1 in \cref{tab:pctl_labels}).
On the other hand, for naive monolithic model checking, it is impossible to extract this probability because it runs out of memory with a model that contains $31,502,736$ states and $51,6547,296$ transitions.
However, at some point, our model checking method is also limited by the size of the induced DTMC and runs out of memory~\citep{DBLP:conf/setta/Gross22}.

\paragraph{How many TMARL agents can our method handle?}
We perform this experiment in the MABP environment with multiple agents because, in this environment, it is straightforward to show how our method performs with different numbers of TMARL agents.
To evaluate our method, we train multiple agents using TMARL to play the MABP game with different numbers of agents.
We then compare the performance of our method to naive monolithic model checking, and evaluate the scalability of both methods as we increase the number of agents in the game.

\begin{figure}[tbp]
  \centering
   {\epsfig{file = 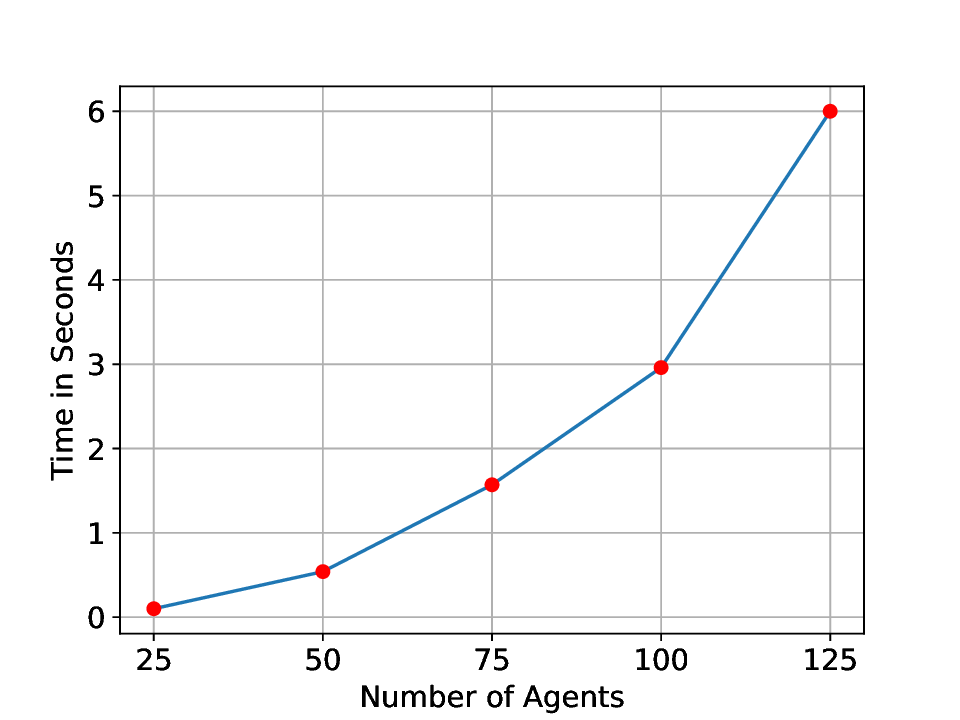, width = 8.2cm}}
  \caption{The diagram shows the time it takes to build a state for a TMARL system as the number of agents in the system increases. The curve in the diagram indicates that the time it takes to build a state increases exponentially as the number of agents increases.}
  \label{fig:time_per_state}
\end{figure}

Naive monolithic model checking is unable to verify ($P^{max}(F\text{ }lost_1)$) $24$ agents in the MABP environment due to memory constraints. The largest possible MDP that can be checked using monolithic model checking contains $23$ agents and has $100,663,296$ states and $247,463,936$ transitions. In contrast, our method allows the model checking up to over $100$ TMARL agents, and we can verify in each of the TMARL systems that agent 1 never uses the riskier bandit (see, for 25 agents, the property query $lost1$ in \cref{tab:pctl_labels}).
This experiment shows, that the limitation of our approach is the action querying time, which increases with the number of agents (see~\cref{fig:time_per_state}).

\paragraph{Do the TMARL agents perform specific game moves?}
In the Pokemon environment, agent 1 uses a heal pot with only 20 HP remaining (see $useHeal20P1$ in \cref{tab:pctl_labels}). This is a reasonable strategy in a late game when the agent is low on HP and needs to restore its HP to avoid being defeated. 
Furthermore, agent 1 wins in the end game with HP=5 and no heal pot left with a probability of $HP5NoHealP1=0.33$, and agent 2 wins with a probability of $HP5NoHealP2=0.66$.
In Tic-Tac-Toe, agent 2 first marks $cell_{10}$, then $cell_{12}$, and finally $cell_{11}$ in a specific order.
In the CC environment, we observe that only the second agent may get knocked out (CC2KO=0.001) and that a collision occurs in $36\%$ of the cases (collision).
Overall, these details show that our method gives insight into policy behaviors in different environments.

\section{\uppercase{Conclusion}}
In this work, we presented an analytical method for model checking TMARL agents.
Our method is based on constructing an induced DTMC from the TMARL system and using probabilistic model checking techniques to verify the behavior of the agents.
We applied our method to multiple environments and found that it is able to accurately verify the behavior of the TMARL agents.
Our method can handle scenarios that can not be verified using naive monolithic model checking methods.
However, at some point, our technique is limited by the size of the induced DTMC and the number of TMARL agents in the system.

In future work, we plan to extend our method to incorporate safe TMARL approaches. This has been previously done in the single agent RL domain~\citep{verifyinloop,DBLP:conf/cav/JothimuruganBBA22}, and we believe it can also be applied to TMARL systems. We also plan to combine our proposed method with interpretable RL techniques~\citep{DBLP:conf/smc/DavoodiK21} to better understand the trained TMARL agents. This could provide valuable insights into the behavior of the agents.

\bibliographystyle{apalike}
{\small
\bibliography{example}}
\appendix

\end{document}